# IMPA-Net: Meteorology-Aware Multi-Scale Attention and Dynamic Loss for Extreme Convective Radar Nowcasting

Haofei Cui[1,2], Guangxin He[1,2], Juanzhen Sun[3], Jingjia Luo[1], Haonan Chen[4], Xiaoran Zhuang[5], Mingxuan Chen[6], Xian Xiao[6]

[1]Key Laboratory of Meteorological Disaster, Ministry of Education (KLME) / International Joint Research Laboratory on Climate and Environment Change (ILCEC) / Collaborative Innovation Center on Forecast and Evaluation of Meteorological Disasters, Nanjing University of Information Science and Technology, Nanjing 210044, China.

[2]Nanchang Meteorological Bureau, Nanchang 330000, China.

[3]National Center for Atmospheric Research, Boulder, CO 80301, USA.

[4]Department of Electrical and Computer Engineering, Colorado State University, Fort Collins, CO 80523, USA.

[5]Jiangsu Meteorological Observatory, Nanjing 210008, China.

[6]Institute of Urban Meteorology, China Meteorological Administration, Beijing 100089, China.

Corresponding author: Guangxin He (002421@nuist.edu.cn)

**Key Points:**

- A meteorology-informed fusion, attention, and loss design improves deterministic radar nowcasting of severe convection
- Asymmetric loss weighting preserves intense echoes and slows the loss of skill with lead time
- The method gives the strongest severe-echo skill among the tested deep learning models while retaining mesoscale structure

**Abstract**

Short-range prediction of convective precipitation from weather radar observations is essential for severe weather warnings, yet deep learning models trained with pixel-wise error metrics inherently produce smooth forecasts that suppress the intense echoes most critical for hazard detection. This underestimation is compounded by insufficient multi-scale feature interaction and suboptimal fusion of heterogeneous geophysical inputs. We present IMPA-Net (Integrated Multi-scale Predictive Attention Network), a deterministic 0–2 hour nowcasting framework that addresses these deficits through coordinated, meteorologically-informed design at the input, architecture, and training levels. A parameter-free Spatial Mixer reorganizes heterogeneous input channels at the mesoscale-γ neighborhood (~2 km) via deterministic channel permutation, providing a structured cross-field prior. An integrated multi-scale predictive attention module serves as the spatiotemporal translator, capturing dynamics from mesoscale-β to mesoscale-γ scales. A Meteorologically-Aware Dynamic Loss employs three-level asymmetric weighting—adapting across training epochs, storm intensity, and forecast lead time—to counteract regression-to-the-mean. Evaluated against seven baselines spanning extrapolation, recurrent, convolutional, and attention-based paradigms on a multi-source radar dataset over eastern China, IMPA-Net raises the Heidke Skill Score at ≥45 dBZ from 0.049 (SimVP baseline) to 0.143 under matched input and training settings. Relative to pySTEPS, IMPA-Net provides a more balanced trade-off between severe-event detection and false-alarm control. Spectral analysis confirms preserved energy across mesoscale bands where competing methods exhibit progressive smoothing. These improvements are demonstrated within a single domain and convective regime; generalizability to regions with different orographic or climatic characteristics remains to be tested.

**Plain Language Summary**

Weather radar continuously scans the atmosphere and produces reflectivity images that indicate where precipitation is occurring and how intense it is. Forecasting how these images will change over the next one to two hours is important for severe weather warnings, but many artificial intelligence models make the forecasts too smooth and weaken the strongest storm echoes. We developed a new deterministic nowcasting model, IMPA-Net, to reduce this problem. The model combines radar observations with terrain and surface rainfall information, analyzes storm evolution at several spatial scales, and uses a training loss that penalizes missed strong echoes more than small overestimates. We tested the model on convective-season radar data from eastern China and compared it with seven benchmark methods. Relative to the other deep learning models, IMPA-Net better preserves intense echoes, loses skill more slowly with lead time, and retains more realistic spatial structure across storm scales. Compared with an extrapolation method, it offers a more balanced trade-off between severe-echo detection and false alarms. These results suggest that meteorology-informed model design can improve short-range radar nowcasting, although broader testing is still needed to confirm how well the method transfers to other regions and climates.

**1 Introduction**

Convective-scale radar nowcasting—predicting the evolution of radar reflectivity fields over 0–2 hour lead times—underpins severe weather warnings and short-range flood preparedness (Ravuri et al., 2021; Zhang et al., 2023). Because the critical decision window for convective hazards is often less than two hours, even modest improvements in nowcast skill can yield disproportionate benefits for emergency response. Operational nowcasting has traditionally relied on radar-echo extrapolation techniques and high-resolution numerical weather prediction (NWP) models. Extrapolation methods such as optical flow are computationally efficient but assume persistence of motion vectors and cannot represent convective initiation or decay (Ayzel et al., 2020). Convection-permitting NWP models can capture these processes through explicit dynamics, yet they require substantial computational resources, suffer from spin-up delays, and remain sensitive to initial condition uncertainties at the convective scale (Sun et al., 2014). Data-driven approaches, enabled by rapid advances in deep learning (LeCun et al., 2015), have emerged as a promising complement to both paradigms, as highlighted by recent calls for tighter integration of machine learning into Earth system prediction workflows (National Academies, 2025), but the task poses persistent challenges. Extreme convective events are rare relative to weak or quiescent conditions, creating highly imbalanced training distributions that bias models toward underestimating high-intensity echoes (Shi et al., 2017; Xu et al., 2024). At the same time, convective systems exhibit tightly coupled dynamics across the mesoscale spectrum—the Meso-$\beta$ organization of squall lines (20–200 km) interacts with Meso-$\gamma$ scale processes such as convective initiation and cell-scale decay (2–20 km)—rendering simple extrapolation insufficient (Sønderby et al., 2020). These difficulties are compounded by progressive lead-time degradation inherent to deterministic forecasts, in which spatial structures blur and intensity peaks erode as the forecast horizon extends, largely because pixel-average loss functions penalize sharp, high-amplitude predictions (Olivetti & Messori, 2024).

Deep learning has advanced radar nowcasting along several complementary modeling pathways. Recurrent architectures such as ConvLSTM (Shi et al., 2015) and PredRNN (Wang et al., 2017; Wang et al., 2022) encode temporal dynamics through sequential hidden states (Hochreiter & Schmidhuber, 1997) but are susceptible to error accumulation over extended horizons. Non-recurrent encoder–translator–decoder frameworks, exemplified by SimVP (Gao et al., 2022) and TAU (Tan et al., 2023), alleviate these issues through purely convolutional or attention-based translators (Vaswani et al., 2017; Dosovitskiy et al., 2021), yet their general-purpose components provide no dedicated mechanism for preserving extreme-intensity features. Domain-specific attention architectures such as SmaAt-UNet (Trebing et al., 2021) and Rainformer (Bai et al., 2022) have incorporated lightweight attention mechanisms tailored to radar fields, improving spatial detail retention, though they do not explicitly address extreme-value underestimation. Generative models, most notably NowcastNet (Zhang et al., 2023), have demonstrated improved perceptual realism, but may introduce hallucinated structures and present optimization challenges in operational settings. On the training objective side, the shortcomings of standard pixel-wise losses such as Mean Squared Error (MSE) have been widely recognized, and alternatives including balanced MSE (Shi et al., 2017), focal loss for class imbalance (Lin et al., 2017), dynamically weighted balanced loss (Fernando & Tsokos, 2022), threshold-aware penalties (Ko et al., 2022; Han et al., 2023), asymmetric extreme-value losses (Xu et al., 2024), and plotting-position-based losses (Xu, L. et al., 2024) have each addressed specific facets of the imbalance problem. Taken together, these advances leave three gaps that motivate the present work: the dynamic orchestration of multiple loss components across training stages and forecast horizons remains largely unexplored; the integration of auxiliary geophysical variables—terrain, low-level wind fields, surface observations—into nowcasting architectures has received limited attention beyond simple channel concatenation (Kim et al., 2024); and no existing deterministic framework jointly addresses input fusion, architectural design, and loss formulation in a coordinated manner for extreme convective radar nowcasting.

With this work, we present IMPA-Net (Integrated Multi-scale Predictive Attention Network), a deterministic radar nowcasting framework in which the input representation, network architecture, and training objective are jointly designed with meteorological considerations. A parameter-free Spatial Mixer encodes local terrain–radar coupling at the Meso-γ scale (~2 km) through fixed geometric permutation of heterogeneous input channels, without introducing trainable parameters. The IMPA module serves as the spatiotemporal translator, combining multi-scale analysis, global self-attention, learnable intensity calibration, and detail restoration to capture cross-scale dynamics while counteracting extreme-value attenuation. The Meteorologically-Aware Dynamic Loss (MAD-Loss) integrates asymmetric extreme-value emphasis, structural similarity, gradient preservation, and temporal consistency under a three-level dynamic weighting strategy that adapts across epochs, storm intensity, and forecast lead time. The framework focuses on the deterministic modeling paradigm as a complement to generative approaches. Unlike physics-informed neural network approaches that embed governing equations as soft constraints (Raissi et al., 2019), IMPA-Net incorporates meteorological knowledge through structured input priors and domain-aware training objectives rather than explicit physical laws.

We evaluate IMPA-Net on the convective seasons (April–September) of 2019–2021 using S-band radar composite reflectivity over Jiangsu Province, China. Through systematic ablation experiments, we demonstrate that the three components yield synergistic improvements, with the integrated framework achieving a 190% relative improvement in Heidke Skill Score for severe convection (≥45 dBZ) over the SimVP baseline (HSS from 0.049 to 0.143). Spectral analysis further confirms that IMPA-Net maintains energy consistency across the Meso-β and Meso-γ scales better than the comparison models.

The remainder of this paper is organized as follows. Section 2 describes the study area, dataset, and task formulation. Section 3 details the methodology, including the Spatial Mixer, IMPA module, MAD-Loss, and the baseline and training configuration. Section 4 presents results covering quantitative comparison, ablation experiments, spectral analysis, and case studies. Section 5 discusses the capabilities and limitations of the framework and concludes the study.

## 2 Materials and Methods

### 2.1 Study Area and Dataset

The study domain covers Jiangsu Province and its surrounding area in eastern China (116.0°E–121.6°E, 30.0°N–34.8°N). The dataset spans the convective seasons (April–September) of 2019–2021 and was provided by the Jiangsu Provincial Meteorological Bureau through the 2022 Jiangsu Meteorological AI Algorithm Challenge. The radar data underwent quality control and mosaic stitching from multiple S-band weather radars across the province before being mapped onto a uniform Cartesian grid of 480 × 560 pixels at 0.01° (~1 km) spatial resolution and 6-minute temporal resolution.

Each sample consists of a 40-frame sequence covering 4 hours, where the first 20 frames (0–114 min) serve as model input and the subsequent 20 frames (120–234 min) constitute the prediction target. We adopt the competition's predefined split, which provides separate training and test sets containing 23,793 and 2,585 sequences, respectively. All results reported in this study are evaluated exclusively on the held-out test set.

2.2 Input Variables and Their Limitations

The model ingests four co-registered input channels at forecast initiation:

(a) Radar composite reflectivity (0–70 dBZ), derived from the S-band radar network as a Constant Altitude Plan Position Indicator (CAPPI) product at 3 km altitude. This channel serves as both the primary input and the sole prediction target. For operational context, reflectivity thresholds of 35 dBZ and 45 dBZ correspond approximately to moderate rainfall (~5 mm h$^{-1}$) and severe convective precipitation (~20 mm h$^{-1}$), respectively, under typical mid-latitude Z–R relationships.

(b) Surface precipitation rate (0–10 mm per 6 min), derived by interpolating 6-minute accumulated precipitation from Automatic Weather Station (AWS) observations across Jiangsu and surrounding areas onto the uniform grid. This field provides an independent, gauge-based constraint on precipitation intensity that complements the radar-derived reflectivity.

(c) Topographic elevation, derived from the SRTM15+ global relief dataset. Although Jiangsu is predominantly flat, localized elevation gradients in the southwestern sector and coastal transitions provide spatial heterogeneity that may modulate low-level convergence and convective triggering.

(d) Along-slope vertical wind component (W_along), a static climatological field computed from the 1988–2018 mean ERA5 reanalysis winds at 850 hPa:

$$W_{along} = u\sin(\theta) + v\cos(\theta)$$

where u and v are the climatological zonal and meridional wind components, and θ is the terrain aspect angle derived from the SRTM15+ elevation data. This variable quantifies the long-term mean upslope (anabatic) or downslope (katabatic) airflow tendency and is intended to encode a persistent orographic forcing prior rather than transient atmospheric dynamics. It remains constant across all time steps within each sequence. We emphasize that this field represents a static climatological background and does not capture event-specific low-level wind variability.

Channels (c) and (d) are time-invariant auxiliary fields that remain constant throughout the 40-frame sequence. The use of static rather than time-varying environmental variables constrains the model's ability to represent transient dynamical forcing; this trade-off is a deliberate design choice that avoids the need for real-time reanalysis ingestion while still providing a terrain-informed spatial prior.

2.3 Task Definition

The nowcasting task is formulated as a deterministic sequence-to-sequence mapping. Given an input tensor $X \in \mathbb{R}^{T_{in} \times C \times H \times W}$ with $T_{in} = 20$ frames, $C = 4$ channels, $H = 480$ rows, and $W = 560$ columns, the model predicts an output tensor $\widehat{Y} \in \mathbb{R}^{T_{out} \times 1 \times H \times W}$ with $T_{out} = 20$ frames of single-channel radar reflectivity. The forecast horizon spans 6 to 120 minutes at 6-minute intervals. The prediction target is radar reflectivity in dBZ; no Z-R conversion to precipitation rate is applied, and all evaluation metrics operate directly in reflectivity space.

# 3 Methodology

## 3.1 Overall Architecture

IMPA-Net adopts an Encoder-Translator-Decoder paradigm (Figure 1), building on the encoder–decoder design principle (Ronneberger et al., 2015) with residual connections (He et al., 2016). The input sequence first passes through the Spatial Mixer (Section 3.2), which fuses the four-channel input into a five-channel representation by concatenating the original radar channel with four geometrically mixed channels. A convolutional encoder then progressively downsamples the spatial dimensions while increasing channel depth, producing latent features $Z \in \mathbb{R}^{B \times T \times C' \times H' \times W'}$. The IMPA module (Section 3.3) serves as the translator, evolving these latent features through a coordinated four-stage process. A symmetric convolutional decoder reconstructs the high-resolution predicted reflectivity sequence $\hat{Y} \in \mathbb{R}^{B \times T_{out} \times 1 \times H \times W}$. The entire framework is trained end-to-end under the MAD-Loss function (Section 3.4).

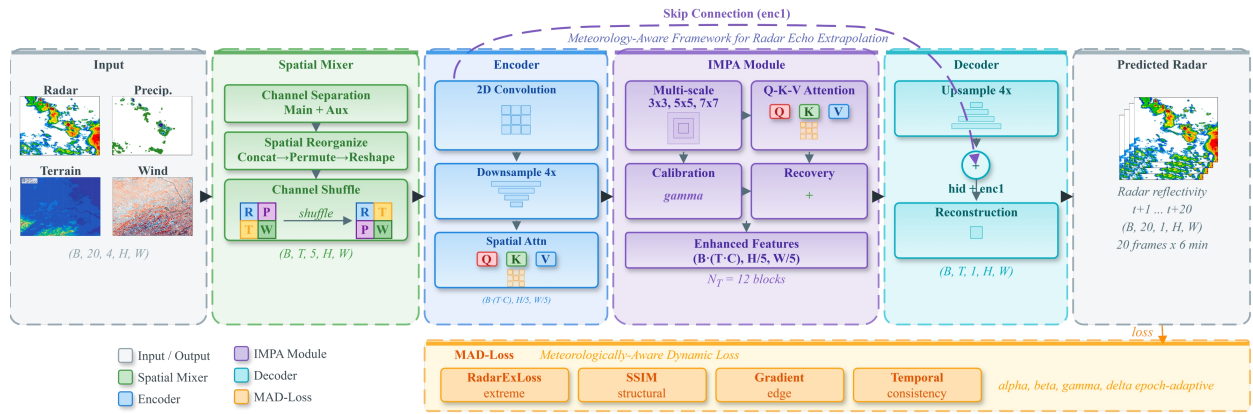

**Figure 1.** Overview of the IMPA-Net framework. Four heterogeneous input channels (radar reflectivity, surface precipitation, topographic elevation, and along-slope wind) are fused by the parameter-free Spatial Mixer into a five-channel representation. A convolutional encoder maps the fused input into latent features, which are processed by N_T = 12 IMPA blocks comprising multi-scale depthwise convolution, global self-attention, learnable intensity calibration, and residual detail recovery. A symmetric decoder with skip connections reconstructs the 20-frame predicted reflectivity sequence (120 min at 6-min intervals). The framework is trained end-to-end under the MAD-Loss, which combines four complementary objectives with epoch-adaptive weights. Tensor dimensions are annotated at key stages.

## 3.2 Spatial Mixer

Standard multi-source fusion approaches, such as channel concatenation, treat distinct geophysical fields as independent channels, placing the full burden of learning inter-field relationships on subsequent network layers. The Spatial Mixer addresses this by encoding local cross-field interactions at the input level through a deterministic, parameter-free geometric permutation.

Given an input tensor $X \in \mathbb{R}^{B \times T \times C \times H \times W}$ with C = 4 heterogeneous channels (radar reflectivity, surface precipitation, topographic elevation, along-slope wind), the spatial dimensions (H, W) are partitioned into non-overlapping $s \times s$ cells, where $s = \lceil \sqrt{C} \rceil$ ($s =$

2 for $C = 4$). Within each $2 \times 2$ cell, the pixel values from the C source channels are spatially rearranged according to a fixed permutation $\sigma_k$ to produce C new mixed channels. Formally, for mixed channel $k \in \{0, ..., C-1\}$:

$$X_{mixed}^{(k)}[2i + p, 2j + q] = X^{(\sigma_k(p,q))}[2i + p, 2j + q], \quad p, q \in \{0,1\}$$

where $\sigma_k$ is a fixed bijection mapping each position (p, q) within the cell to a source channel index. For example, in the first mixed channel (k = 0), the top-left pixel originates from channel 0 (radar), the top-right from channel 1 (precipitation), the bottom-left from channel 2 (terrain), and the bottom-right from channel 3 (wind). Each mixed channel thus interleaves information from all four source fields at the spatial resolution of a $2 \times 2$ cell. At the dataset's 0.01° (~1 km) grid spacing, this corresponds to a ~2 km neighborhood, comparable to the Meso-$\gamma$ scale at which terrain–precipitation interactions are most pronounced.

Although the permutation rule is fixed, the inherent spatial heterogeneity of the input fields—time-varying radar patterns versus static terrain—ensures that the resulting mixed channels are globally distinct from one another. To retain the global spatial continuity of the radar field, which is essential for tracking storm motion, the original radar channel is preserved and concatenated with the four mixed channels along the channel dimension:

$$X_{input} = concat[X^{(0)}, X_{mixed}^{(0)}, X_{mixed}^{(1)}, X_{mixed}^{(2)}, X_{mixed}^{(3)}], \quad dim = channel$$

yielding a 5-channel input $X_{input} \in \mathbb{R}^{B \times T \times (C+1) \times H \times W}$ to the encoder.

The Spatial Mixer introduces zero trainable parameters and acts as a deterministic reorganization operator that enhances local cross-field visibility for downstream layers. It does not model physical processes explicitly; rather, it provides a structured prior that allows the encoder to perceive co-located terrain–radar–precipitation relationships without relying on learned fusion weights. When the number of auxiliary channels C differs from 4, the cell size can be adjusted to $s = \lceil \sqrt{C} \rceil$ with appropriate zero-padding for non-square cases. At coarser grid resolutions, larger cell sizes may be warranted to maintain a comparable physical neighborhood scale. The complete operation is illustrated in Figure 2.

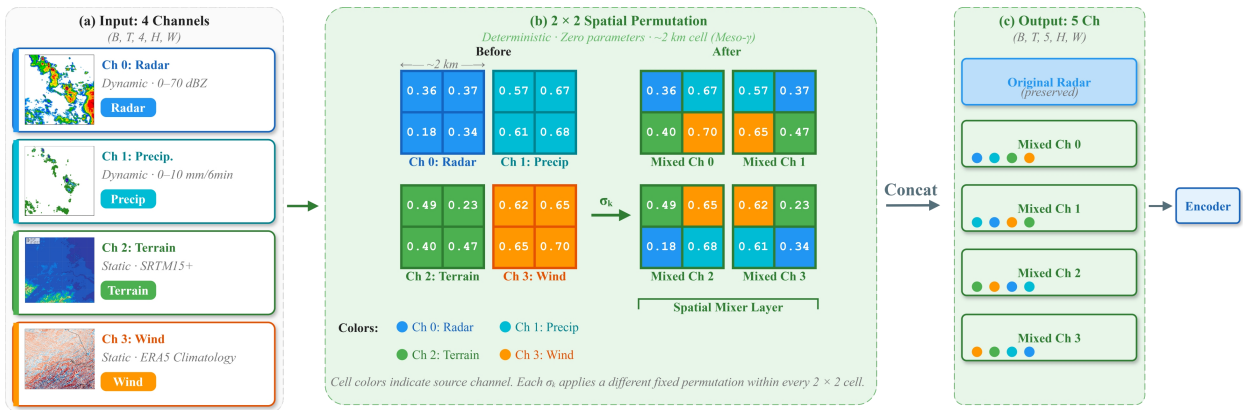

**Figure 2.** Schematic of the Spatial Mixer operation. (a) Four heterogeneous input channels—radar reflectivity, surface precipitation, topographic elevation, and along-slope wind—each shown with a representative data sample. (b) Within every non-overlapping $2 \times 2$ cell (~2 km at 0.01° resolution), four fixed permutations rearrange pixel values across channels, producing

four mixed channels in which each cell interleaves information from all source fields. Cell colors in the mixed channels indicate the source channel of each value. (c) The original radar channel is preserved and concatenated with the four mixed channels, yielding a five-channel input to the encoder $\sigma_0$–$\sigma_3$.

### 3.3 IMPA Module

The IMPA (Integrated Multi-scale Predictive Attention) module replaces the generic convolutional translator used in SimVP-family architectures (Gao et al., 2022) with a four-stage processing pipeline designed for extreme-value spatiotemporal prediction (Figure 3).

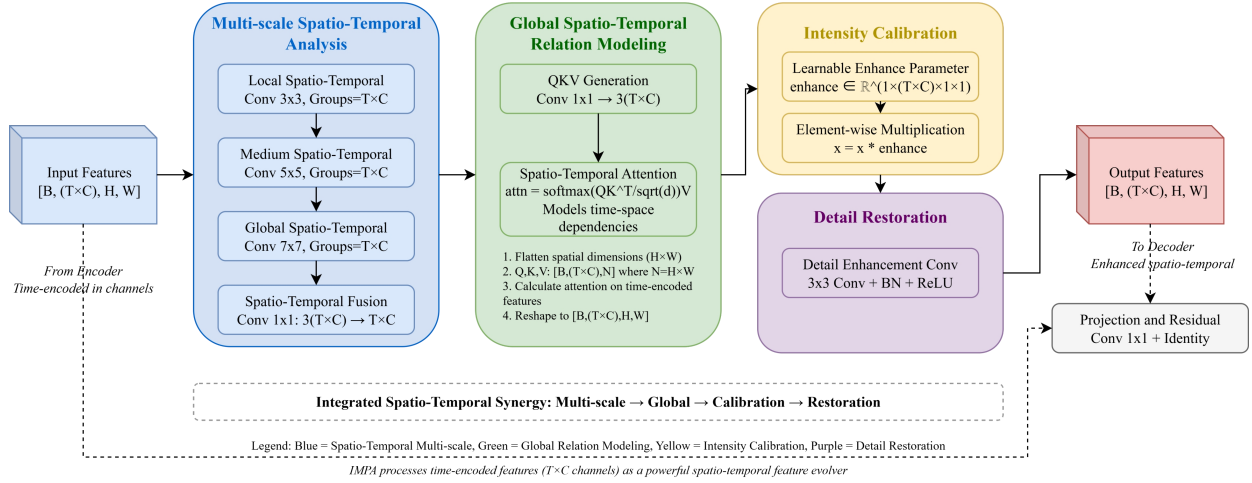

**Figure 3.** Integrated processing pipeline within the IMPA module, combining multi-scale analysis, global attention, intensity calibration, and detail restoration. Temporally-encoded features first pass through parallel multi-scale convolutional branches (Stage 1). Global spatial dependencies are then modeled via self-attention (Stage 2). A learnable parameter performs channel-wise intensity calibration to enhance extreme values (Stage 3). Finally, a residual block restores fine-scale details (Stage 4). The entire process is integrated with residual connections to facilitate training.

The module first reshapes the temporally ordered latent features $Z \in \mathbb{R}^{B \times T \times C' \times H' \times W'}$ by folding time into the channel dimension to obtain $Z_{encoded} \in \mathbb{R}^{B \times (T \cdot C') \times H' \times W'}$, enabling efficient 2D spatial operators to implicitly model joint spatiotemporal dynamics.

Stage 1: Multi-scale analysis. Three parallel depthwise convolutional branches with kernel sizes $3 \times 3$, $5 \times 5$, and $7 \times 7$ extract features at different spatial scales. Group convolution (groups = $T \cdot C'$) maintains temporal separation during spatial extraction. A learned $1 \times 1$ convolution fuses the branch outputs into a scale-aware representation $F_{multi}$.

Stage 2: Global relation modeling. Following the self-attention mechanism (Vaswani et al., 2017), global self-attention is applied across the spatial domain ($H' \times W'$) of $F_{multi}$. Query, key, and value projections are computed, and attention weights capture dependencies between all spatial locations, yielding context-aware features $F_{global}$.

Stage 3: Intensity calibration. A learnable channel-wise scaling parameter $\gamma \in \mathbb{R}^{1 \times (T \cdot C') \times 1 \times 1}$ is applied element-wise: $F_{calibrated} = \gamma \odot F_{global}$. Optimized jointly with the MAD-

Loss, γ adaptively amplifies feature channels associated with high-intensity events. Unlike Feature-wise Linear Modulation (FiLM; Perez et al., 2018), which applies affine transformations conditioned on an external input (γ = f(condition), β = f(condition)), our calibration parameter is globally learned during training without external conditioning. This simpler formulation avoids the need for an auxiliary conditioning pathway while still allowing the network, guided by the asymmetric MAD-Loss penalties, to learn channel-specific amplification factors that counteract extreme-value attenuation.

Stage 4: Detail restoration. A residual convolutional block (He et al., 2016) (Conv 3 × 3 → BN → ReLU) refines $F_{calibrated}$. An additive residual connection $F_{restored} = F_{calibrated} + F_{detail}$ preserves fine-scale structures such as storm boundaries.

A global residual connection $Z_{encoded} \leftarrow Z_{encoded} + F_{restored}$ promotes stable training. The final output is obtained after a projection layer and reshaping back to [B, T, C', H', W'] for the decoder.

3.4 Meteorologically-Aware Dynamic Loss (MAD-Loss)

MAD-Loss integrates four complementary objectives under a three-level dynamic weighting strategy. The total loss takes the form:

$$L_{MAD} = \alpha(e) \cdot \tilde{L}_{ext} + \beta(e) \cdot L_{ssim} + \gamma(e) \cdot L_{grad} + \delta(e) \cdot L_{temp}$$

where $\tilde{L}_{ext}$ is the per-frame weighted extreme loss (detailed below), and the remaining three terms-structural similarity $L_{ssim} = 1 - SSIM(P_{norm}, T_{norm})$, spatial gradient $L_{grad} = ||\nabla P - \nabla T||_1$, and temporal consistency $L_{temp} = ||(P_t - P_{t-1}) - (T_t - T_{t-1})||_1$ are each computed over the full output sequence.

Radar Extreme Loss. For each frame t, $L_{ext}^{(t)}$ augments standard MSE with an asymmetric penalty within extreme intensity zones. Underestimation is penalized with coefficient $\alpha_{under} = 1.2$ versus $\alpha_{over} = 1.0$ for overestimation, encoding the operational priority of avoiding missed severe events. For later frames ($t \geq T/2$), $\alpha_{under}$ increases to 1.6 and the extreme threshold is lowered from 35 to 30 dBZ to further counteract lead-time degradation.

Three-level dynamic weighting. (1) Epoch-adaptive component weights $\alpha(e), \beta(e), \gamma(e), \delta(e)$ are defined by sigmoid schedules and normalized to sum to unity at each epoch:

$$\alpha_{raw}(e) = 0.4 + 0.2 \cdot (1 - \sigma(e/10)), \quad \beta_{raw}(e) = 0.25 + 0.2 \cdot \sigma(e/5)$$

$$\gamma_{raw}(e) = 0.15 + 0.1 \cdot (1 - \sigma(e/15)), \quad \delta_{raw}(e) = 0.15 + 0.1 \cdot \sigma(e/10)$$

This schedule prioritizes extreme-value emphasis and gradient sharpness early in training, then gradually shifts toward structural and temporal objectives. (2) A storm-aware factor $w_{storm}(t) = 1 + (2.0 + 1.2 \cdot I_{t \geq T/2}) \cdot r_t$ amplifies the loss for frames with a high fraction $r_t$ of pixels exceeding 40 dBZ. (3) Piecewise-linear temporal weights λ(t) increase from 1.0 → 0.9 over the first half of the forecast horizon and from 1.2 → 2.0 over the second half, rescaled so that $\sum_t \lambda(t) = T$. These two per-frame weights apply exclusively to the extreme loss branch:

$$\tilde{L}_{ext} = (1/T) \cdot \sum_t \lambda(t) \cdot w_{storm}(t) \cdot L_{ext}^{(t)}$$

concentrating storm-aware and lead-time-aware adaptation where it matters most, while the remaining three objectives provide stable sequence-level regularization.

**Relation to prior work.** MAD-Loss shares the asymmetric penalty concept with balanced MSE (Shi et al., 2017), focal loss (Lin et al., 2017), dynamically weighted balanced loss (Fernando & Tsokos, 2022), and ExLoss (Xu et al., 2024). It extends these approaches by combining four objectives and employing a three-level dynamic hierarchy. The epoch-adaptive schedule is predefined rather than learned; exploring learned weighting is a direction for future work.

### 3.5 Baselines and Training Configuration

**Evaluation metrics.** Forecast skill is assessed using threshold-based contingency metrics computed at reflectivity levels of 35 dBZ (moderate) and 45 dBZ (severe convection): Critical Success Index (CSI), Heidke Skill Score (HSS), Probability of Detection (POD), False Alarm Ratio (FAR), and Frequency Bias. All threshold metrics are reported as 20-frame averages unless stated otherwise. Structural fidelity is evaluated using the Structural Similarity Index (SSIM), Learned Perceptual Image Patch Similarity (LPIPS), and Radially Averaged Power Spectral Density (RAPSD). To summarize spectral differences within each scale band, we compute the log-spectral error (LSE), defined as the root-mean-square difference between the $\log_{10}$-transformed RAPSD of the forecast and the observation over the wavenumbers corresponding to the Meso-β (20–200 km) and Meso-γ (2–20 km) bands, respectively.

**Baseline models.** IMPA-Net is compared against seven reference methods spanning extrapolation, recurrent, convolutional, and attention-based paradigms: ConvLSTM-based MIM (Wang et al., 2019), SimVP (Gao et al., 2022), PredRNNv2 (Wang et al., 2022), TAU (Tan et al., 2023), the domain-specific architectures RainNet (Ayzel et al., 2020) and SmaAt-UNet (Trebing et al., 2021), and the Lagrangian extrapolation system pySTEPS (Pulkkinen et al., 2019). All deep learning baselines are trained on the same dataset with identical input configurations and optimized using Mean Squared Error (MSE) loss; pySTEPS uses default parameter settings without training. This setup provides a controlled comparison that isolates the effect of architecture and, in the case of IMPA-Net, the combined effect of architecture and loss design.

**Training configuration.** IMPA-Net is implemented in PyTorch and trained end-to-end for 40 epochs using the MAD-Loss function described above. We employ the AdamW optimizer with an initial learning rate of $5 \times 10^{-4}$ and a cosine annealing scheduler with a batch size of 2 due to the large spatial dimensions (480 × 560) of the input frames. All experiments are conducted on NVIDIA A100 GPUs (40 GB) with CUDA 12.2.

## 4 Results

### 4.1 Forecast Skill Across Intensity Thresholds

Table 1 summarizes forecast skill across three reflectivity thresholds for all eight models. At ≥20 dBZ (light precipitation), CSI values are broadly similar across models, ranging from 0.459 to 0.537, with RainNet (0.537) and PredRNNv2 (0.525) performing best. IMPA-Net attains a CSI of 0.473 at this threshold, indicating competitive light-precipitation detection.

**Table 1.** Quantitative Comparison of Precipitation Forecast Skill Across Models

| Category | Model | Overall | | ≥20 dBZ | ≥35 dBZ | | ≥45 dBZ | | | |
| --- | --- | --- | --- | --- | --- | --- | --- | --- | --- | --- |
| | | MAE↓ | SSIM↑ | CSI↑ | CSI↑ | POD↑ | CSI↑ | POD↑ | FAR↓ | Bias |
| Extrapolation | pySTEPS | 0.052 | **0.644** | 0.459 | 0.268 | 0.386 | **0.093** | **0.154** | 0.815 | 0.832 |
| General ST-DL | MIM | 0.044 | 0.546 | 0.516 | 0.230 | 0.245 | 0.043 | 0.044 | **0.234** | 0.058 |
| | SimVP | 0.045 | 0.548 | 0.520 | 0.270 | 0.315 | 0.053 | 0.057 | 0.360 | 0.095 |
| | PredRNNv2 | **0.043** | 0.559 | 0.525 | 0.267 | 0.301 | 0.058 | 0.062 | 0.429 | 0.111 |
| | TAU | 0.047 | 0.476 | 0.503 | 0.197 | 0.211 | 0.022 | 0.023 | 0.216 | 0.030 |
| Domain-specific DL | RainNet | **0.043** | 0.606 | **0.537** | 0.246 | 0.269 | 0.048 | 0.050 | 0.274 | 0.069 |
| | SmaAt-UNet | 0.044 | 0.603 | 0.508 | 0.214 | 0.233 | 0.055 | 0.060 | 0.407 | 0.119 |
| Proposed | **IMPA-Net** | 0.045 | 0.605 | 0.473 | **0.277** | **0.379** | 0.084 | 0.128 | 0.551 | 0.450 |

**Note.** All metrics are 20-frame (6–120 min) averages over 2585 test sequences. Best values are in bold; second best are underlined. All DL baselines use MSE loss; IMPA-Net uses MAD-Loss. Frequency Bias = (H + F)/(H + M); Bias > 1 indicates over-forecasting, Bias < 1 indicates under-forecasting, and ideal = 1.0. At 45 dBZ, pySTEPS achieves the highest CSI and POD but at a false-alarm ratio of 81.5%, whereas IMPA-Net provides a substantially lower FAR (55.1%) with the highest frequency bias (0.450) among the DL models.

Differences become more pronounced at higher thresholds. At ≥35 dBZ (moderate convection), IMPA-Net achieves the highest CSI (0.277) and POD (0.379), followed by SimVP in CSI (0.270) and pySTEPS in POD (0.386). At ≥45 dBZ (severe convection), IMPA-Net retains the second-highest CSI (0.084) and POD (0.128) among all methods, behind only pySTEPS (CSI = 0.093, POD = 0.154). This threshold-dependent pattern, with smaller differences at low intensity and larger relative gains at moderate-to-severe intensity, indicates that IMPA-Net places more forecast skill on high-intensity echo features.

pySTEPS requires separate interpretation because it is a Lagrangian extrapolation method rather than a learned forecast model. It attains the highest SSIM (0.644), as well as the highest ≥45 dBZ POD and frequency bias, by advecting existing echoes along the estimated motion field without modeling intensity evolution (Pulkkinen et al., 2019). This mechanism preserves spatial structure in the short term and retains severe echoes that are already present, but it cannot represent initiation, intensification, or decay. Its FAR of 81.5% at ≥45 dBZ indicates the resulting trade-off: echoes that have weakened or dissipated in the observations are frequently retained in the forecast. By contrast, the deep learning models tend to attenuate echo intensity over time, which lowers FAR but often suppresses severe events too strongly, as reflected in their low frequency biases (<0.12). IMPA-Net lies between these two behaviors. Its bias of 0.450, the highest among the deep learning models, indicates that it preserves a substantially larger fraction of severe echoes than the other deep learning baselines while avoiding the indiscriminate persistence of extrapolation.

Despite this emphasis on severe-event detection, IMPA-Net maintains competitive overall forecast quality. Its MAE (0.045) is within 5% of the best-performing model (PredRNNv2: 0.043), and its SSIM (0.605) ranks third overall, just below pySTEPS (0.644) and RainNet (0.606). These results indicate that the improved detection of moderate-to-severe echoes does not come at the expense of overall forecast accuracy or structural fidelity. The component-level basis for this behavior is examined in Section 4.3, and its process-level implications are revisited in Section 4.5.

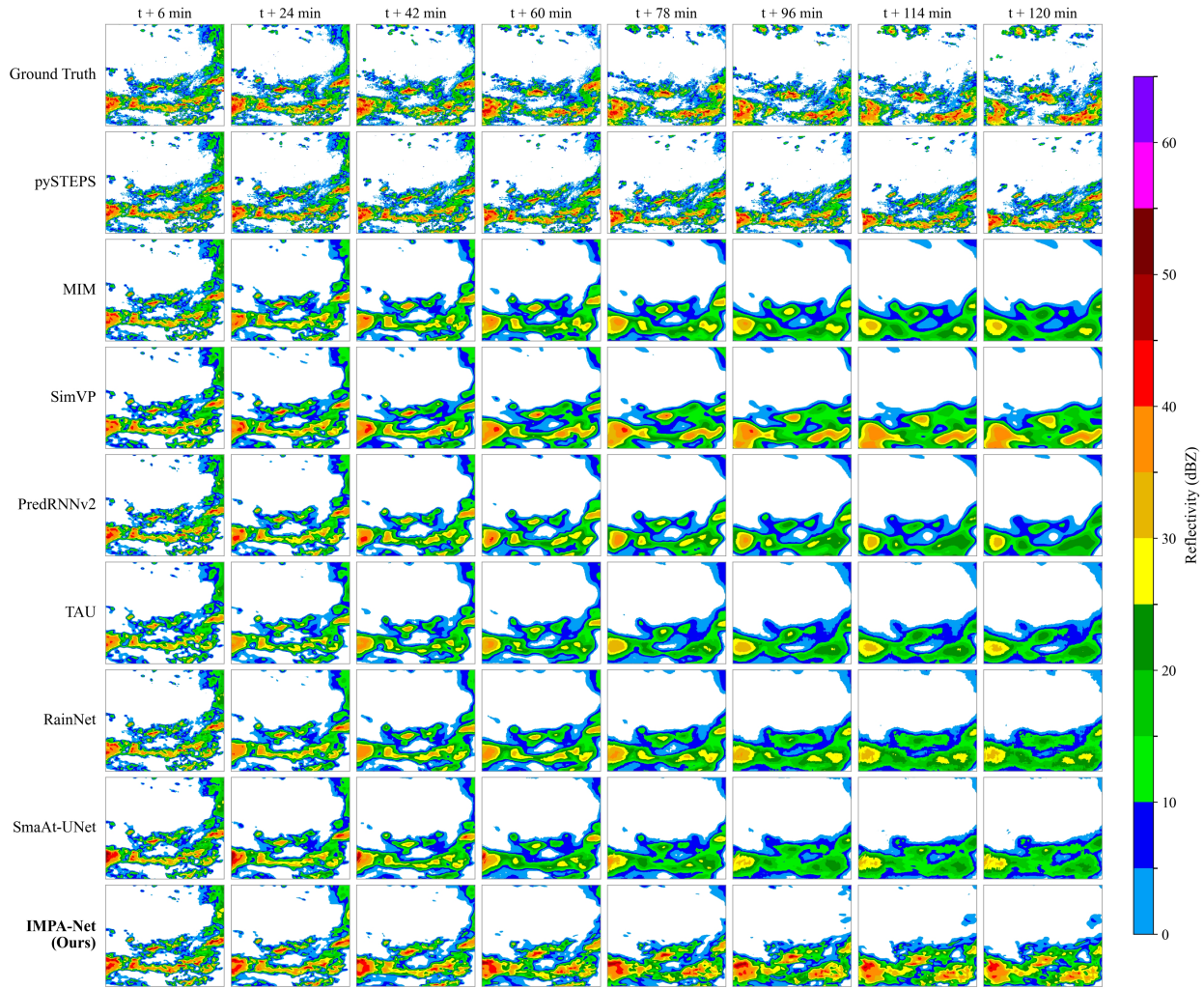

**Figure 4.** Multi-model radar reflectivity forecasts for a representative severe-convection case (Case 1593). Each row shows one model's predicted reflectivity fields at eight lead times (t + 6, 24, 42, 60, 78, 96, 114, and 120 min); the top row is the ground truth. Most deep-learning baselines (MIM, SimVP, PredRNNv2, TAU) progressively smooth the echo field, with the high-reflectivity core largely vanishing by t+60 min. pySTEPS preserves echo structure longer but generates spatially displaced false alarms at later lead times. IMPA-Net (bottom row) maintains the convective core and intensity gradients more faithfully throughout the forecast window.

Figure 4 illustrates these contrasting behaviors for a representative test case (Case 1593). Most DL baselines (MIM, SimVP, PredRNNv2, TAU) progressively smooth the echo field, with the high-reflectivity core largely vanishing by T+60 min—consistent with their declining CSI@45 dBZ beyond the first hour (Section 4.2). pySTEPS preserves echo structure longer but propagates features that have already dissipated in the observation, generating spatially displaced false alarms coherent with its elevated FAR. In contrast, IMPA-Net maintains the convective core through the forecast window more faithfully, reproducing the observed intensity gradients and organizational structure at lead times where the other models have smoothed to near-uniform fields.

## 4.2 Lead-Time Degradation of Forecast Skill

Figure 5 examines the temporal evolution of severe-convection forecast skill over the 6–120 min forecast horizon. All models exhibit monotonic degradation with increasing lead time, consistent with the intrinsic predictability limits of meso-γ scale processes (Surcel et al., 2015). However, the rate and shape of this degradation differ substantially between methods.

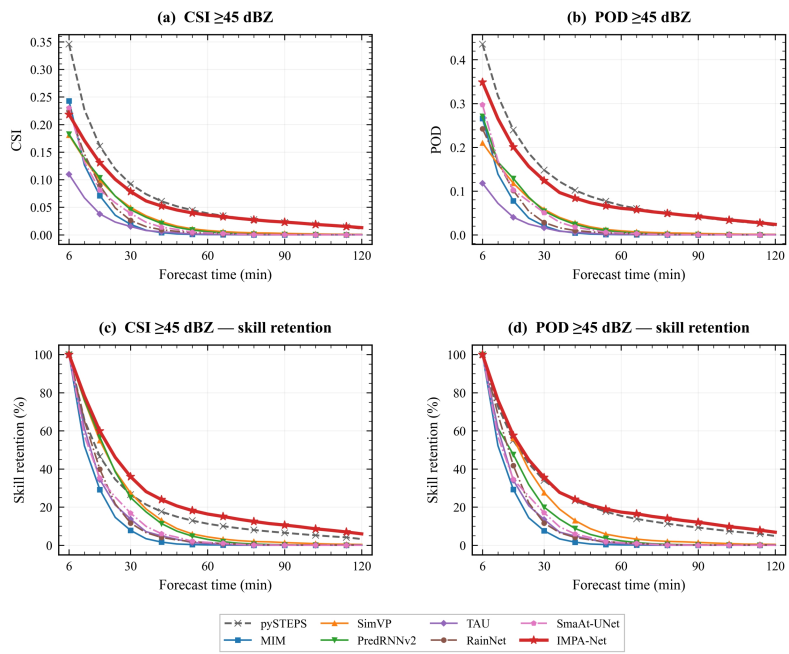

**Figure 5.** Lead-time degradation of severe-convection forecast skill (≥45 dBZ) over the 6–120 min forecast horizon for eight models. (a) CSI and (b) POD as functions of forecast lead time; IMPA-Net (red) maintains nonzero skill through the full 120-min window, whereas most deep-learning baselines decay to near-zero by 48–72 min. (c) CSI skill retention and (d) POD skill retention, defined as each metric normalized by its T+6 min value and expressed as a percentage, isolating the intrinsic degradation rate from differences in initial skill magnitude. IMPA-Net exhibits the slowest decay among all methods in both metrics.

Panels (a) and (b) present the absolute CSI and POD at ≥45 dBZ as a function of lead time. Among the deep learning models, most baselines, including MIM, PredRNNv2, TAU, RainNet, and SmaAt-UNet, decay to near-zero CSI within approximately 48–72 min (Figure 5a). By T+90 min, their CSI values are effectively zero (<0.001), indicating that severe-event detection skill is largely lost well before the end of the forecast window. IMPA-Net, by contrast, retains nonzero CSI through the full 120-min horizon, decreasing from 0.218 at T+6 min to 0.013 at T+120 min. Although this represents a 94% reduction, it remains about an order of magnitude larger than the corresponding values of the other deep learning baselines. The POD curves in Figure 5b show the same pattern: IMPA-Net remains the highest among the deep learning models at every lead time, declining from 0.349 to 0.024, whereas the other baselines fall below 0.005 by T+60 min.

To isolate the effect of initial skill differences on temporal robustness, panels (c) and (d) show skill retention, defined as each metric normalized by its own T+6 min value and expressed as a percentage. This normalization removes differences in initial magnitude and emphasizes the intrinsic degradation rate. Under this view, IMPA-Net exhibits the slowest decay among all methods for both CSI (Figure 5c) and POD (Figure 5d) at ≥45 dBZ. At T+120 min, IMPA-Net retains approximately 6.0% of its initial CSI and 6.8% of its initial POD, whereas the next-best deep learning models (SimVP and PredRNNv2) retain less than 0.4%. Even pySTEPS, despite its higher initial absolute values, retains only 3.2% of its CSI, less than half of the IMPA-Net retention rate, and this residual skill is increasingly accompanied by false alarms, as discussed in Section 4.1.

IMPA-Net degrades more slowly than the other models, so its long-range advantage reflects a lower decay rate rather than just higher initial skill. This temporal robustness helps explain why IMPA-Net remains useful beyond the first hour. The component-level factors contributing to this behavior are examined in the ablation analysis (Section 4.3).

4.3 Component Contributions and Synergy

To quantify how each proposed component contributes to the overall forecast skill, we conduct a systematic ablation study using eight configurations obtained by progressively removing the Spatial Mixer, the IMPA module, and MAD-Loss (Figure 6). The reference baseline is the original SimVP backbone, an encoder–translator–decoder architecture with GSTA as the translator, trained with standard MSE loss and without any of the three proposed components.

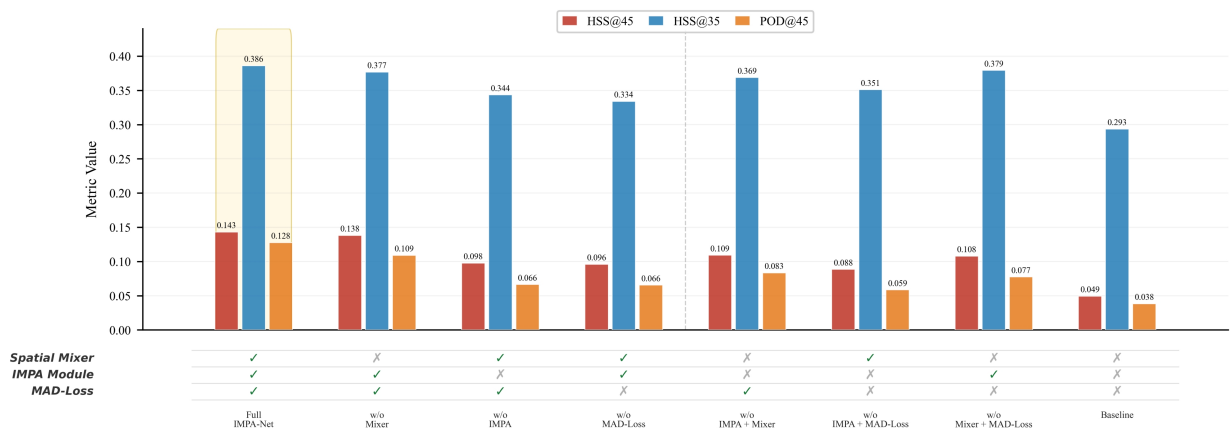

**Figure 6.** Ablation study of the three proposed components—Spatial Mixer, IMPA module, and MAD-Loss—evaluated on HSS at ≥45 dBZ (red), HSS at ≥35 dBZ (blue), and POD at ≥45 dBZ (orange). Bars show the 20-frame averaged metric values for eight configurations: the full IMPA-Net (highlighted), three single-component removals, three double-component removals, and the bare SimVP baseline. The check/cross grid beneath each configuration indicates which components are retained. A vertical dashed line separates the single-removal and double-removal groups. All experiments share the same dataset, training schedule, and backbone architecture.

Among single-component additions to the baseline, MAD-Loss yields the largest isolated gain in HSS at ≥45 dBZ (+0.060) and POD at ≥45 dBZ (+0.045), consistent with its role in shifting the optimization emphasis toward severe-event detection. The IMPA module produces the largest single-component improvement in HSS at ≥35 dBZ (+0.086), reflecting its contribution to preserving moderate-to-severe echo structure through multi-scale, intensity-aware processing. The Spatial Mixer alone provides moderate gains in both HSS at ≥35 dBZ (+0.058) and HSS at ≥45 dBZ (+0.039), although this benefit is smaller than that of the other two components, which is consistent with the relatively weak terrain heterogeneity of the Jiangsu domain noted in Section 2.

The removal analysis reinforces these findings. Removing MAD-Loss from the full model causes the largest degradation in HSS at ≥45 dBZ (0.143 → 0.096, −33%) and POD at ≥45 dBZ (0.128 → 0.066, −49%), highlighting its central role in severe-event detection. Removing the IMPA module produces a comparable HSS at ≥45 dBZ decline (0.143 → 0.098, −32%), whereas removing the Spatial Mixer has the smallest effect (0.143 → 0.138, −3.5%). Notably, the full model attains a POD of 0.128 at ≥45 dBZ, which is a 233% improvement over the baseline value of 0.038. Taken together, these results show that the strongest severe-event skill is achieved only when the three components operate jointly.

These ablation experiments are conducted within the same study domain and dataset. The relative contribution of each component, particularly that of the Spatial Mixer, may differ in regions with more complex terrain, such as the mountainous periphery of the Sichuan Basin or the Central Mountain Range of Taiwan.

### 4.4 Spatial Structure Fidelity

In addition to threshold-based and pixel-wise metrics, the radially averaged power spectral density (RAPSD) quantifies how well forecast fields retain the observed spatial variability across scales. Figure 7 compares the observed spectrum with all baselines at five lead times, with emphasis on the Meso-β band (20–200 km), associated with mesoscale convective organization, and the Meso-γ band (2–20 km), associated with storm-core structure.

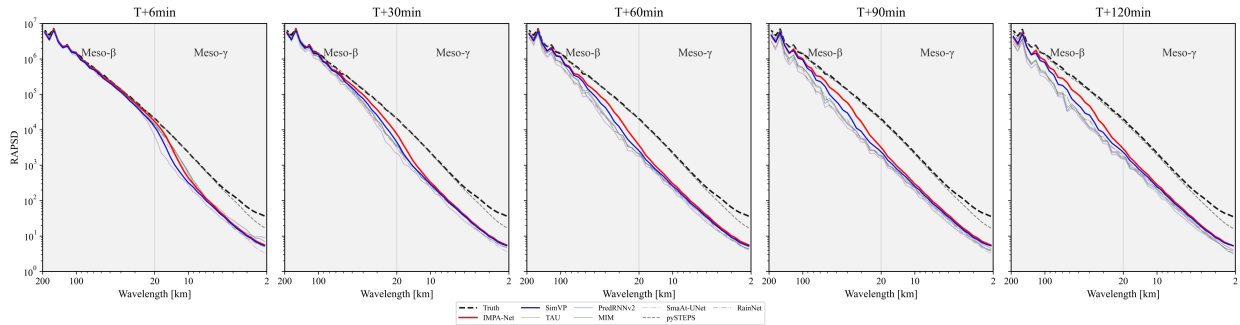

**Figure 7.** Radially averaged power spectral density (RAPSD) of observed and forecast reflectivity fields at five lead times (T+6, T+30, T+60, T+90, and T+120 min), averaged over all test sequences. The vertical gray line at 20 km separates the Meso-β band (20–200 km) from the Meso-γ band (2–20 km). The observed spectrum (black dashed) serves as the reference. Model-specific differences are discussed in Section 4.4.

At T+6 and T+30 min, all models reproduce the observed spectral slope reasonably well, and the largest departures are confined to the shortest resolved wavelengths. With increasing lead time, however, most deep learning models exhibit a systematic reduction in spectral power in both bands, consistent with progressive smoothing of forecast structure. By T+120 min, IMPA-

Net remains the closest deep learning model to the observed spectrum. Its Meso-β LSE is 0.39, compared with 0.54 for SimVP and 0.74–0.86 for RainNet, SmaAt-UNet, MIM, PredRNNv2, and TAU. In the Meso-γ band, IMPA-Net again yields the smallest LSE among the deep learning models (0.85, versus 0.90 for SimVP and 0.97–1.06 for the remaining baselines). These results indicate that IMPA-Net retains more realistic spatial variability at both the organizational and storm-core scales, particularly at longer lead times.

pySTEPS exhibits the highest spectral fidelity at every lead time, with a T+120 Meso-β LSE of 0.05, because it advects the most recent observed field rather than explicitly predicting intensity evolution. This spectral advantage should therefore not be interpreted as evidence of uniformly better forecast performance. As shown in Section 4.1, pySTEPS attains the highest SSIM and the strongest ≥45 dBZ CSI, POD, and HSS, but these scores are accompanied by very large false-alarm rates (FAR = 0.544 at ≥35 dBZ and 0.815 at ≥45 dBZ) and the lowest CSI at ≥20 dBZ (0.459). Taken together, the RAPSD and threshold-based metrics indicate that pySTEPS is especially effective at retaining existing echo patterns, whereas IMPA-Net provides the most favorable balance among the deep learning models between spatial-structure preservation and skill in representing echo evolution.

### 4.5 Convective Lifecycle Representation

The preceding subsections evaluate IMPA-Net from a statistical perspective. Here we examine whether the model also captures process-level aspects of convective evolution that are only partially reflected in aggregate metrics. Figure 8 focuses on Case 1715 and compares three representative reference methods: pySTEPS as an extrapolation-based nowcast, RainNet as a domain-specific deep learning baseline, and IMPA-Net as the proposed model. In the ground truth, an arc-shaped convective rainband translates toward the southeast over the 120-min forecast window while maintaining a coherent band-scale organization. The sequence exhibits three identifiable phases: (i) an early mature stage with embedded intense cores (≥45 dBZ), (ii) weakening and structural reorganization of the southwestern sector between approximately T+42 and T+60 min, and (iii) renewed development near the southeastern end of the band, which becomes visible around T+60 min and intensifies further after T+78 min. This combination of band-scale translation and internally varying intensity presents a more stringent test than simple echo advection.

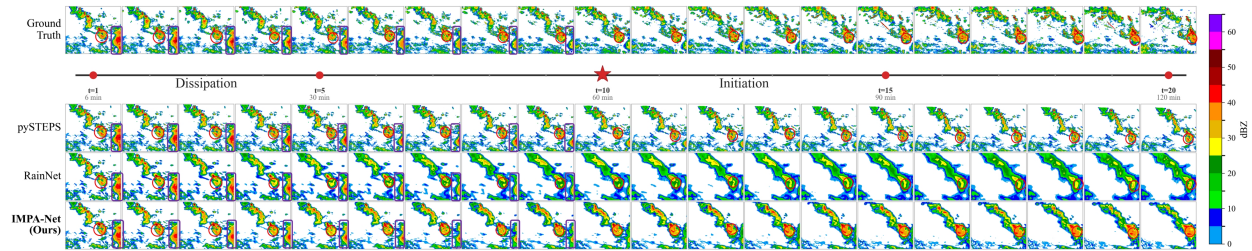

**Figure 8.** Convective lifecycle representation for Case 1715. Four rows show the 20-frame (6–120 min) reflectivity evolution for Ground Truth, pySTEPS, RainNet, and IMPA-Net. The ground-truth sequence exhibits mature-stage intense cores, dissipation on the eastern flank (purple boxes), and renewed initiation of a central cell (red circles) after approximately T+42 min. Differences in how each model handles this intensity redistribution are discussed in Section 4.5.

The three comparison methods in Figure 8 exhibit distinct behaviors consistent with their quantitative statistics. pySTEPS preserves the displacement and band-scale morphology through mid-range lead times, but it primarily advects existing echoes forward and does not represent the observed intensity adjustment: strong cores persist after they have weakened in the observations, consistent with the high FAR reported in Table 1. RainNet retains the broad rainband envelope but progressively loses the internal intensity contrast and fine-scale structure; by the later lead times, the ≥45 dBZ cores are strongly muted, consistent with its rapid skill decay in Section 4.2. IMPA-Net better reproduces both the southeastward translation of the band and the subsequent redistribution of intensity, retaining sharper embedded cores through the transition period and more closely matching the later redevelopment near the southeastern end of the system.

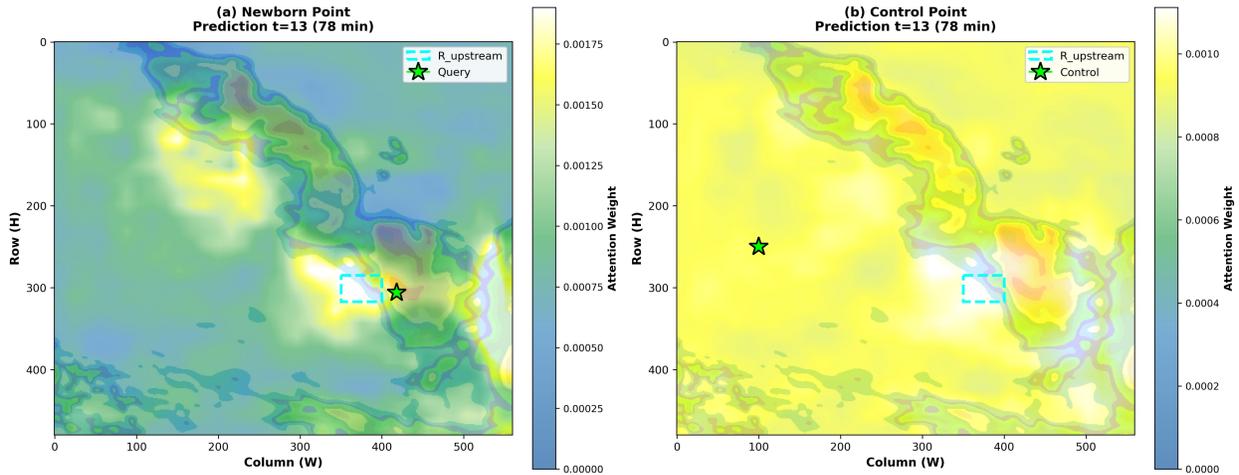

**Figure 9.** Spatial attention attribution from Layer 2 of the IMPA module for Case 1715 at T+78 min, overlaid on the IMPA-Net forecast field. (a) Attention pattern when querying a redevelopment point near the southeastern end of the convective band (green star); attention concentrates on an upstream convective portion approximately 30–45 km away (2.08× enhancement relative to the spatial average, cyan dashed box). (b) Attention pattern for a control point in a quiescent region, showing diffuse, non-specific attention (1.16× enhancement). The contrast between the two patterns suggests that the IMPA module learns spatially selective dependencies aligned with the mesoscale organization of the convective band.

Spatial attention attribution from Layer 2 of the IMPA module provides complementary evidence for this interpretation (Figure 9). When querying the redevelopment region near the southeastern end of the band at T+78 min, the attention concentrates on an upstream convective portion of the same band approximately 30–45 km away (2.08× enhancement relative to the spatial average), whereas a control point in a quiescent region shows diffuse, non-specific attention (1.16× enhancement). This upstream-downstream dependence is broadly consistent with the evolving mesoscale organization of the convective band, although it should be interpreted as a single-case illustration rather than a systematic attribution analysis.

## 5 Discussion and Conclusions

### 5.1 Capabilities and Limitations

Section 4 supports three conclusions about the capabilities of IMPA-Net within the evaluated deterministic nowcasting setting. First, the framework improves the detection of moderate-to-severe convective echoes relative to the deep learning baselines, with the clearest gains at ≥35 and ≥45 dBZ. Second, these gains are not confined to a single verification perspective: they are reflected in threshold-based skill scores, slower lead-time degradation, improved retention of spectral power across the Meso-β and Meso-γ bands, and more faithful reproduction of the evolving rainband structure in the representative case studies. Third, the ablation analysis indicates that these improvements do not arise from a single dominant design choice alone, but from the coordinated interaction of the loss function, the translator architecture, and the structured input representation.

At the same time, the results also clarify what the present framework can and cannot justify. IMPA-Net should be interpreted as a meteorology-aware deterministic radar-nowcasting model rather than as an explicit physical simulator of convective dynamics. The auxiliary terrain and along-slope wind fields provide a static environmental prior, and the attention analysis suggests that the model learns meaningful non-local statistical dependencies. However, the current inputs do not contain time-varying dynamical fields that would allow direct attribution to specific mechanisms such as cold-pool evolution, gravity-wave triggering, or low-level convergence. Accordingly, the case-study and attention results are best interpreted as evidence that the model captures process-relevant spatial dependencies, not that it explicitly resolves the underlying atmospheric dynamics.

The framework also exhibits a clear operational trade-off. Relative to other deep learning baselines, IMPA-Net allocates more forecast skill to severe echoes and retains a larger fraction of intense cores at longer lead times. This behavior is desirable for high-impact nowcasting, but it is accompanied by a higher false-alarm burden than the more strongly smoothed deep learning baselines. The comparison with pySTEPS further illustrates that no single method dominates all aspects of performance: extrapolation preserves existing structure effectively, but its skill at high thresholds is accompanied by substantial false alarms and limited ability to represent intensity adjustment. In this sense, IMPA-Net's contribution is not to eliminate the underlying trade-off, but to move the balance toward a more useful compromise between severe-event detection, spatial-structure retention, and process-level realism.

Two additional limitations remain important. The first concerns the static nature of the environmental priors. In the present formulation, topography and climatological along-slope flow can constrain the spatial context of convection, but they cannot replace event-specific dynamic environmental information. Incorporating time-varying wind, thermodynamic, or reanalysis fields is a natural next step for testing whether the framework can better represent transient forcing. The second concerns generalizability. All experiments are conducted over Jiangsu Province during the convective seasons of 2019–2021. Because Jiangsu is characterized by relatively modest terrain variability and a single observational system, the present results do not establish that the same gains will transfer unchanged to regions with different orographic, climatic, or radar-network characteristics. Broader validation across regions, weather regimes, and years will therefore be necessary to assess the portability of the proposed design. These observations suggest a preliminary hierarchy of design contributions: in domains with weak terrain heterogeneity, the loss formulation and translator architecture are the primary drivers of severe-echo skill, while the structured environmental prior may become more influential in regions with stronger orographic forcing—a hypothesis that warrants cross-regional testing.

5.2 Summary

In summary, IMPA-Net demonstrates that a coordinated design across input fusion, translator architecture, and loss formulation can improve deterministic convective radar nowcasting within the evaluated 0–120 min window. Across quantitative metrics, spectral analysis, and representative case studies, the model more effectively preserves intense echoes, maintains spatial variability across scales, and reproduces key aspects of rainband evolution than the deep learning baselines considered here. These gains should be interpreted within the bounds of the current evidence: the model is meteorology-aware, not a substitute for explicit dynamical modeling, and its behavior remains conditioned by static priors and single-region evaluation. Still, the results show that meteorology-aware design can improve both severe-event skill and structural fidelity in deterministic radar nowcasting, and they provide a starting point for future work incorporating dynamic environmental forcing and broader out-of-domain validation.

**Inclusion in Global Research**

This research was conducted collaboratively across institutions in China and the United States. All authors contributed to the study design, analysis, or manuscript preparation based on their respective expertise, regardless of geographic location.

**Acknowledgments**

This work was supported by the State Key Laboratory of Climate System Prediction and Risk Management (CPRM) initiative project (Grant No. CPRM-2025-NUIST-012); China Meteorological Administration Capability Enhancement Joint Research Program (24NLTSQ015); Sichuan Science and Technology Program (No. 2025YFNH0006); Water Conservancy Science and Technology Project of Jiangsu Province (No. 2025017); China Meteorological Administration Innovation and Development Program (CXFZ2023J008); Scientific Research Project of Jiangsu Meteorological Service (KM202520); the Open Fund Project for Heavy Rain (BYKJ2024Q23); and the Bohai Rim Regional Meteorological Science and Technology Collaborative Innovation Fund project (QYXM202409).

**Open Research**

The IMPA-Net source code, analysis scripts, and a representative subset of example radar composite reflectivity observations used in this study are publicly available via GitHub (https://github.com/Nanming666/IMPA-Net) and are permanently archived on Zenodo (https://doi.org/10.5281/zenodo.19495345).

The full, continuous radar composite reflectivity dataset is proprietary and was provided by the Jiangsu Provincial Meteorological Bureau. A detailed description and prior application of this identical restricted dataset can be found in Zhuang et al. (2023). Due to strict Chinese government regulations regarding meteorological data security and the Bureau's internal policies, these raw data cannot be distributed publicly. Researchers wishing to access the full dataset must apply directly to the Jiangsu Provincial Meteorological Bureau, subject to official approval and appropriate data use agreements.

Topographic elevation data were obtained from the SRTM15+ global relief dataset (https://topex.ucsd.edu/WWW_html/srtm15_plus.html). ERA5 reanalysis data used to derive the climatological along-slope wind component are available from the Copernicus Climate Data Store (https://cds.climate.copernicus.eu).

**Conflict of Interest Disclosure**

The authors declare there are no conflicts of interest for this manuscript.